\documentclass{article}
\usepackage{ijcai26}

\usepackage{times}
\usepackage{soul}
\usepackage{url}
\usepackage[hidelinks]{hyperref}
\usepackage[utf8]{inputenc}
\usepackage[small]{caption}
\usepackage{graphicx}
\usepackage{amsmath}
\usepackage{amsfonts}
\usepackage{amsthm}
\usepackage{booktabs}
\usepackage{algorithm}
\usepackage{algorithmic}
\usepackage[switch]{lineno}
\usepackage{xcolor}
\usepackage{adjustbox}
\usepackage{array}
\usepackage{multirow}
\usepackage{amssymb}
\usepackage[hang,flushmargin]{footmisc}

\def\eg{\emph{e.g.}} 
\def\ie{\emph{i.e.}} 
 
\def\etc{\emph{etc.}}

\title{Progressive Reasoning with Primitive Correction\\ for Compositional Zero-Shot Learning}

\author{
Ziyi Chen\and
Haoyan Shi\and
Sunhan Xu\And
Congyan Lang\\
\affiliations
School of Computer Science and Technology, Beijing Jiaotong University\\Key Laboratory of Big Data \& Artificial Intelligence in Transportation (Ministry of Education)
\emails
\{zychen, hy\_shi, sunhan\_xu, cylang\}@bjtu.edu.cn
}

\begin{document}

\maketitle


\begin{abstract}
Compositional Zero-Shot Learning (CZSL) aims to combine known attributes and objects as primitives for recognizing previously unseen attribute-object pairs. 
Prior works either predict attributes and objects independently, missing their strong contextual dependency, or use unidirectional conditional modeling (\eg, object-guided attribute prediction), which is prone to error propagation. 
We propose PRPC, a \textbf{P}rogressive \textbf{R}easoning framework with \textbf{P}rimitive \textbf{C}orrection, which explicitly models the bidirectional dependency between attributes and objects via step-wise inference. 
PRPC performs mutual correction of primitives to suppress prediction errors in earlier steps.
Specifically, we formulate CZSL as structured, Q\&A-style Chain-of-Thought reasoning process and constrain the MLLM to follow predefined semantic steps to generate intermediate decisions. 
To further enhance the reliability and logical consistency of intermediate reasoning, we introduce reinforcement learning post-training with a GRPO-based objective, providing step-level rewards aligned with the progressive inference procedure. 
Extensive experiments on three CZSL benchmarks demonstrate that PRPC achieves state-of-the-art performance, validating the effectiveness of progressive reasoning and bidirectional correction for robust compositional generalization.
\end{abstract}

\section{Introduction}

\begin{figure}[ht]
    \centering
    \includegraphics[width=0.49\textwidth]{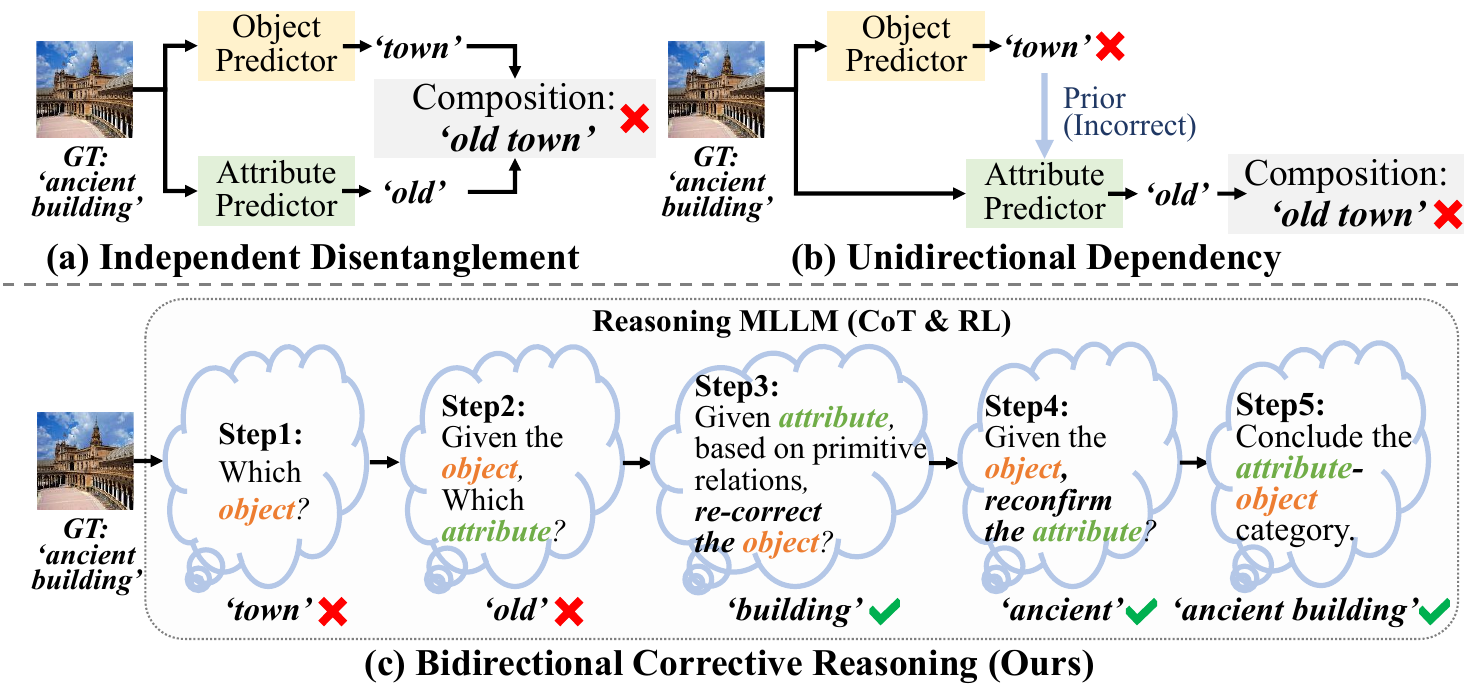}
    \caption{Comparison of compositional learning paradigms. (a) Early approaches disentangle attributes and objects independently, ignoring crucial contextual correlations. (b) Recent methods introduce unidirectional dependency, typically using object predictions as priors. However, incorrect object prediction inevitably misleads attribute inference. (c) Our proposed framework introduces a bidirectional corrective reasoning mechanism, in which attribute predictions correct initial object hypotheses and vice versa, reducing error propagation through iterative refinement.}
    \label{fig:intro}
\end{figure}

Humans naturally have the ability to generalize different learned concepts (\eg, objects and attributes), to form new compositions. For example, with images of `yellow bird' and `red flower' in mind, when an unseen image of `yellow flower' emerges, humans can effortlessly comprehend such a new composite concept. Similarly, in language-based computer vision tasks (\eg, image or video captioning~\cite{video_captioning} and navigation~\cite{navigation}) with insufficient training samples, recognizing novel composite concepts is crucial for visual understanding. Considering this, \textbf{c}ompositional \textbf{z}ero-\textbf{s}hot \textbf{l}earning (CZSL), aiming to classify images into previously unseen attribute-object pair labels by learning primitives (\ie, attributes or objects) from images with known pair labels, has gained increasing attention. 


Early approaches typically treat primitive recognition as two independent tasks~\cite{CVPR2023ADE,CSP}. 
By disentangling features into parallel predictors, these methods learn to identify attributes and objects separately. 
While effective for understanding primitives in unseen samples, this paradigm often overlooks the intrinsic interaction between attributes and objects, which is commonly referred to as contextuality, as shown in Figure~\ref{fig:intro} (a). 
The visual representation of an attribute is strongly correlated with its paired object; for instance, a `wet' road looks significantly different from a `wet' cat. 
Neglecting these correlations fundamentally limits generalization performance across varying contexts.

To address this, recent methods~\cite{CVPR2023CANet,ICCV2023hierarchical} have moved beyond pure disentanglement by introducing conditional dependencies, typically using object predictions as priors to guide attribute estimation, as illustrated in Figure~\ref{fig:intro} (b). 
While this progressive strategy models primitive relations, it suffers from a critical drawback: error propagation. 
In a unidirectional pipeline, an incorrect object prediction inevitably misleads the subsequent attribute inference, degrading overall accuracy.

In this paper, we propose a \textbf{P}rogressive \textbf{R}easoning framework with \textbf{P}rimitive \textbf{C}orrection, termed \textbf{PRPC}, which explicitly models stepwise dependencies between objects and attributes for compositional zero-shot learning. 
Motivated by the mutually constraining nature of objects and attributes, we move beyond treating object recognition as a one-way prior for attribute inference. 
Instead, we design a bidirectional corrective reasoning mechanism in which attribute reasoning actively refines object recognition, allowing attribute and object predictions to be updated through iterative verification. 
Specifically, we formulate CZSL as a structured multi-step decision process:
(1) predict which object is present; (2) given the object, predict which attribute applies; (3) given the attribute, verify and correct the object by grounding the decision in token-level visual information; (4) given the updated object, verify and correct the attribute; and (5) output the final attribute–object category. 
The object–attribute refinement steps (\ie, Steps 3-4) form a mutual verification and correction loop, which alternates between attribute and object reasoning, mitigates error propagation from earlier decisions, and progressively improves compositional consistency. 

Inspired by the strong, structured reasoning behavior of Chain-of-Thought (CoT) prompting~\cite{CoT}, we implement the above procedure in a Q\&A-style format and constrain a reasoning-capable MLLM (\eg, Qwen-VL~\cite{Qwen}) to produce structured intermediate decisions that follow these predefined semantic steps. 
Furthermore, we introduce an RL-based post-training objective based on the GRPO loss \cite{GRPO}, providing step-level rewards to encourage faithful and logically consistent intermediate decisions throughout the verification loop.

Our main contributions are summarized below: 
\begin{itemize}
\item We propose PRPC, a novel framework that transcends independent primitive prediction by introducing explicit progressive reasoning. We are the first to systematically introduce the reasoning paradigm with CoT to CZSL and build an MLLM-based benchmarking framework.
\item We design a bidirectional corrective reasoning mechanism that utilizes attribute cues to refine object recognition, effectively mitigating the error accumulation inherent in unidirectional pipelines. Additionally, we integrate CoT constraints with GRPO-based reinforcement learning to enhance the reliability of intermediate reasoning steps.
\item Extensive experiments on three benchmark datasets demonstrate that PRPC achieves state-of-the-art performance, validating the effectiveness of our progressive reasoning strategy.
\end{itemize}

\section{Related Works}

\subsection{Compositional Zero-Shot Learning (CZSL)}

Compositional Zero-Shot Learning (CZSL) is a specialized branch of ZSL that classifies novel attribute-object pairs by learning primitives from seen pairs. Existing methods mainly differ in how explicitly they disentangle these primitives.

Early CZSL approaches \textit{do not explicitly disentangle primitives}, treating each attribute–object pair as a holistic concept and learning joint embeddings between images and compositional label~\cite{CVPR2017,ICCV2019-TMN}. 
They perform well on seen pairs but often overfit training co-occurrences, limiting generalization to unseen compositions.

To improve compositionality, later work introduces disentanglement. 
\textit{Textual disentanglement} exploits the attribute–object separability in language, learning contextualized composition representations from primitive embeddings~\cite{ECCV2018,xu2021relation} or retaining primitive semantics alongside composition features~\cite{HPL}, but text alone cannot capture object-dependent visual variations of attributes. 
\textit{Visual feature disentanglement} separates attribute- and object-related cues in image features via augmentation~\cite{SCEN,ICCV2023hierarchical}, prototype anchors~\cite{Nips2021-GCN,qulearning}, or auxiliary objectives encouraging consistency~\cite{CVPR2023ADE,saini2024beyond}.
Recently, \textit{cross-modal disentanglement} adapts CLIP~\cite{CLIP} (\eg, prompting or encoder modifications) to align and disentangle primitives across vision and language, achieving strong results in open-world CZSL~\cite{CSP,Troika,PLID}. 

Different from methods with independent primitive prediction or one-way conditioning, we formulate recognition as progressive bidirectional reasoning between attributes and objects, enabling mutual correction and stronger compositional generalization.

\begin{figure*}[ht]
\centering
\includegraphics[width=1.0\textwidth]{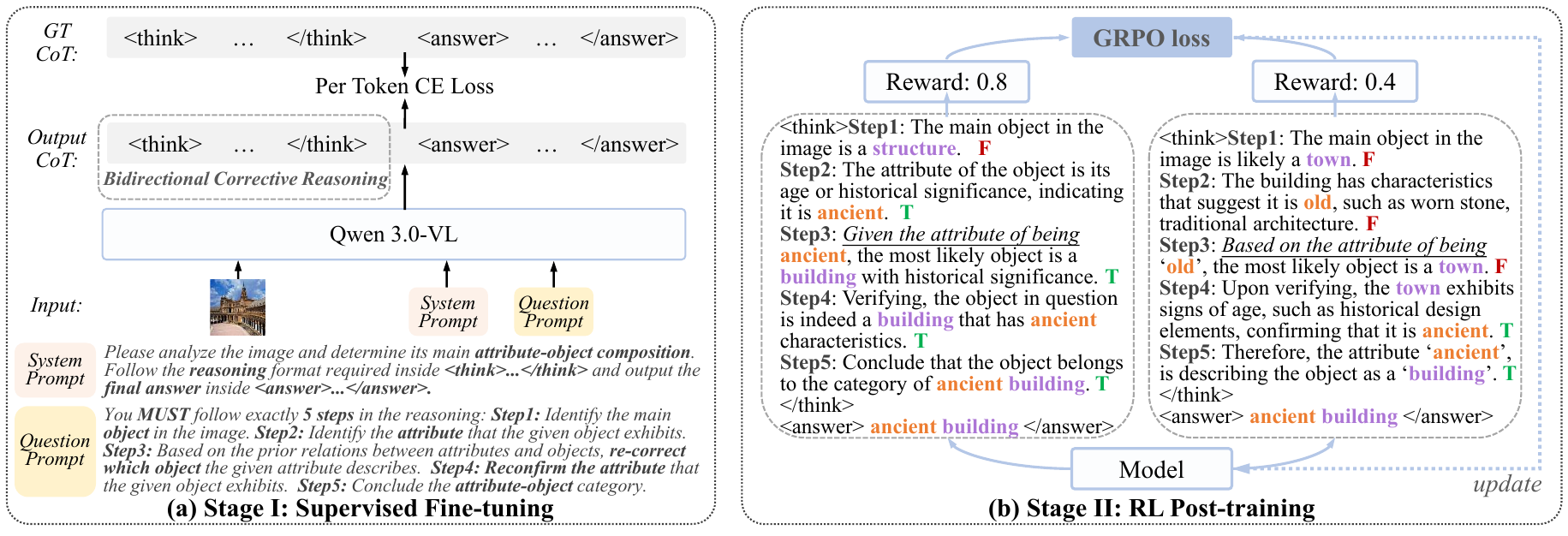}
\caption{Illustration of our framework PRPC with two-stage training. During training, PRPC uses a CoT-style output format to support step-by-step composition reasoning, consisting of two parts, \ie, \texttt{<think>} and \texttt{<answer>}. In \texttt{<think>}, we apply Bidirectional Corrective Reasoning to correct primitive predictions and implement step-wise RL rewards.}
\label{fig:overview}
\end{figure*}

\subsection{MLLM Fine-Tuning and RL Post-Training}

Recent advances in Multimodal Large Language Models (MLLMs) largely come from supervised fine-tuning and reinforcement learning–based post-training. 
Fine-tuning typically adapts pretrained multimodal backbones to downstream tasks by aligning visual and textual representations with task-specific annotations.
Common strategies include instruction tuning on image–text pairs~\cite{LLaVA,MiniGPT-4}, region-level supervision~\cite{MDETR,GLIP}, and cross-modal contrastive objectives~\cite{BLIP}, which improve multimodal grounding, reasoning consistency, and task generalization.

Reinforcement learning post-training has emerged as a complementary paradigm that further refines model behavior beyond supervised signals. 
By optimizing task-level or preference-based rewards, RL post-training enables models to better follow complex instructions. 
In multimodal scenarios, rewards can be defined over textual outputs~\cite{christiano2017deep,ouyang2022training}, visual grounding quality~\cite{yu2024rlhf,yan2024vigor}, or cross-modal consistency~\cite{sun2024aligning}.
By directly optimizing task-level rewards, RL post-training complements fine-tuning and enhances robustness and controllability.


\subsection{Chain-of-Thought (CoT) in Visual Reasoning}

Chain-of-Thought (CoT) prompting improves multi-step reasoning in large language models by eliciting intermediate inference steps, boosting performance on arithmetic, symbolic, and commonsense reasoning tasks~\cite{zhouleast}. 
Follow-up work shows that simple decoding strategies, such as self-consistency, can further strengthen CoT reliability~\cite{yao2023tree}. 
More recently, CoT-style reasoning has been extended to vision–language settings, where models generate stepwise explanations that are encouraged to align with visual evidence (\eg, objects, attributes, and relations), improving interpretability and robustness on multimodal reasoning benchmarks~\cite{LLaVA,InstructBLIP}. 
Several approaches also combine CoT with tool- or program-based visual reasoning or instruction-tuned multimodal LLMs to enable more controllable, compositional grounding~\cite{ViperGPT,VisProg}. 


\section{Our Approach}

\subsection{Problem Formulation}
Let $\mathcal{A}= \{ a_1, a_2, \dots, a_{|\mathcal{A}|}\}$ be an attribute set, $\mathcal{O}= \{ o_1, o_2, \dots, o_{|\mathcal{O}|}\}$ be an object set, and $\mathcal{C} = \mathcal{A} \times \mathcal{O}$ be the composition set, defined as $\mathcal{C}= \{(a,o) \mid a\in \mathcal{A}, o\in \mathcal{O}\}$. 
The composition set $\mathcal{C}$ is divided into two disjoint subsets: the seen composition set $\mathcal{C}_s$ and the unseen composition set $\mathcal{C}_u$, where $\mathcal{C}_s \cap \mathcal{C}_u = \emptyset$. 
The dataset is split into a seen set $\mathcal{T}_{s} = \{(x, c) \mid x \in \mathcal{I}, c \in \mathcal{C}_{s}\}$ and an unseen set $\mathcal{T}_{u}$. 

Traditional approaches seek a mapping function $f: \mathcal{I} \to \mathcal{C}$ that directly outputs the highest-scoring pair. Here, we treat CZSL as a reasoning paradigm. 
We cannot include a candidate label set in the MLLM prompt for two reasons. 
First, it dilutes contextual attention. 
Many short labels are highly similar at the token level, so the MLLM does not compare them one by one but relies on coarse semantic information.
Second, MLLMs are generative models designed to produce the most likely outputs, and thus cannot function as closed-set classifiers like CLIP-style VLMs. 

We therefore formulate MLLM-based generative approaches to predicting compositional labels as \textbf{\textit{open-form compositional prediction}}. 
This learning setting is substantially more challenging than the standard CZSL paradigm.





\subsection{Framework Overview} 

We propose a new CZSL framework named PRPC, as shown in Figure~\ref{fig:overview}, which introduces explicit progressive reasoning to model relations between primitives. 
Motivated by the strong reasoning ability of Chain-of-Thought (CoT), we cast CZSL as a structured, Q\&A-style reasoning procedure. 
Concretely, we guide a reasoning-capable MLLM (\eg, Qwen-VL~\cite{Qwen}) to follow a predefined chain of semantically grounded steps. 
During training, the model outputs in the CoT format, which includes multi-step thinking \texttt{<think>} \dots \texttt{</think>} and the final composition label inside \texttt{<answer>} \dots \texttt{</answer>}. 
For each input instance $x$ (\eg, image + question), the policy generates a sequence $y$ which can be formulated as:
\begin{equation}
y = \big(y^{\text{think}},\, y^{\text{ans}}\big),
\end{equation}
where $y^{\text{think}}$ is enclosed by \texttt{<think>} \dots \texttt{</think>} and $y^{\text{ans}}$ is enclosed by \texttt{<answer>} \dots \texttt{</answer>}. 

We design a bidirectional corrective reasoning mechanism that leverages attribute cues to refine object recognition, enabling our model to accurately capture the interactions between primitives and effectively mitigating the error accumulation inherent in one-way pipelines.

Specifically, training PRPC consists of two stages.  
\textbf{Stage I (SFT)}: Given an image and its ground-truth composition, we generate stepwise CoT targets to supervise the baseline model to follow a template and perform coherent multi-step reasoning.  
\textbf{Stage II (RL)}: We further fine-tune the baseline with a GRPO loss, evaluating each step in the progressive CoT via exact matching and assigning corresponding rewards.

\subsection{Bidirectional Corrective Reasoning}

We disentangle compositional concept recognition into distinct object and attribute predictions across five explicitly defined reasoning steps. 
Since objects typically dominate visual representations~\cite{CVPR2023CANet}, Step1 queries the model for the object category first. 
Step2 follows a strategy similar to prior CZSL works, where the predicted object serves as a prior for attribute inference. 

We subsequently introduce a bidirectional correction mechanism to prevent initial object misclassification from compromising the final result. 
Step3 infers the object conditioned on the attribute, while Step4 reconfirms the attribute for verification. 
Attributes and objects in compositional learning exhibit strong co-occurrence and feasibility constraints. 
Inferring the object from the attribute eliminates implausible candidates and addresses common confusion among visually similar objects. 
The attribute thus functions as a disambiguation cue to validate the initial object prediction and mitigate error propagation.

Finally, an example of our model's output is as follows:

\noindent{\fontfamily{lmtt}\selectfont \textcolor[RGB]{46,117,182}{<think>} \textcolor[RGB]{128,128,128}{Step1:} The main object in the image is a \textcolor[RGB]{84,130,53}{building}. \\ \textcolor[RGB]{128,128,128}{Step2:} The attribute of this object is its age or historical significance, as the label specifies it is `\textcolor[RGB]{191,144,0}{ancient}'. \\ \textcolor[RGB]{128,128,128}{Step3:} Based on the attribute of being \textcolor[RGB]{191,144,0}{ancient}, the most likely object is a \textcolor[RGB]{84,130,53}{building} that has been preserved for its historical and cultural value, such as a monument, temple, or fortress. \\ \textcolor[RGB]{128,128,128}{Step4:} Ancient buildings typically have specific architectural styles, materials, or decay that align with historical periods, confirming this \textcolor[RGB]{84,130,53}{building} has the attribute `\textcolor[RGB]{191,144,0}{ancient}'. \\ \textcolor[RGB]{128,128,128}{Step5:} Conclude that the object with the given attribute is an \textcolor[RGB]{191,144,0}{ancient} \textcolor[RGB]{84,130,53}{building}. \textcolor[RGB]{46,117,182}{</think>}\textcolor[RGB]{46,117,182}{<answer>}\textcolor[RGB]{191,144,0}{ancient} \textcolor[RGB]{84,130,53}{building}\textcolor[RGB]{46,117,182}{</answer>}}

\subsection{Generated CoT Supervision}

Standard training annotations provide only the correct composition $c$, while the full CoT trace sentences are not uniquely determined. 
High-quality supervision is essential for inducing explicit reasoning capabilities in the model. 

To this end, we employ GPT-4o~\cite{GPT-4o} to generate structured reasoning traces that align with our CoT formulation, enabling stepwise supervision of the reasoning process.
We impose structural constraints on the LLM-generated content primarily to mitigate the training and evaluation instability caused by open-form generation. 
By standardizing each step into fixed fields, we ensure that the parsing modules can reliably extract the target strings for primitive prediction. 

For each training sample, we provide GPT-4o with the image and the ground-truth label $c$, instructing it to produce a fluent five-step reasoning trace $y^{\text{think}*}$. 
Notably, we use parsing functions to verify the correctness of the generated CoT ground truth. 
Finally, the composition label is encapsulated within $y^{\text{ans}*}$, which forms the complete ground-truth output $y^*$.


\subsection{Stage I: Supervised Fine-tuning}

We start with supervised fine-tuning (SFT) to establish a reliable initialization, which serves two purposes. 
First, it teaches the model to consistently follow our output specification: a five-step \texttt{<think>} block followed by a concise \texttt{<answer>}. 
Second, it encourages each step to expose primitive intermediate predictions that align with our step-wise evaluation protocol, allowing later stages to score progress at the granularity of individual steps.

We minimize the standard Cross-Entropy (CE) loss over all target tokens:
\begin{equation}
\begin{adjustbox}{max width=\linewidth}
$\begin{aligned}
\mathcal{L}_{\text{SFT}}(\theta) = &-\mathbb{E}_{(x,y)} \Bigg[ \alpha \sum_{i=1}^{|y^{\text{think}}|} \log \pi_\theta(y^{\text{think}}i \mid x, y^{\text{think}}{<i}) \\&+ \beta \sum_{j=1}^{|y^{\text{ans}}|} \log \pi_\theta(y^{\text{ans}}j \mid x, y^{\text{think}}, y^{\text{ans}}{<j}) \Bigg],
\end{aligned}$
\end{adjustbox}
\label{eq:sft}
\end{equation}


After SFT converges, we obtain an initial policy $\pi_{\theta_0}$ that achieves high compliance with the structured output specification and provides a strong starting point for GRPO-based optimization under strict exact-matching rewards.

\subsection{Stage II: RL Post-training}


While supervised fine-tuning enables the model to adhere to the CoT format, its token-level supervision can restrict the diversity of reasoning trajectories and limit generalization beyond the training distribution. 
To overcome the inherent constraints of purely supervised optimization, we further optimize the policy with reinforcement learning to maximize step-wise correctness. 

We face three main challenges: (1) exact-match scoring yields sparse and discontinuous rewards, (2) training is highly sensitive to formatting errors, since a single missing tag can break downstream parsing, and (3) learning from a single sampled trajectory leads to high gradient variance.
To mitigate these issues, we adopt Group Relative Policy Optimization (GRPO)~\cite{GRPO} for post-training. 
GRPO updates the policy by sampling multiple candidate trajectories for each input. 
It selectively reinforces responses using a group-relative baseline to boost task-level rewards, eliminating the need to train a separate critic model.

\noindent\textbf{Structured generation and parsing.}
A deterministic parser extracts the intermediate step strings
\begin{equation}
\big(s_1,s_2,s_3,s_4,s_5\big) = \mathrm{Parse}\big(y^{\text{think}}\big),
\end{equation}
where each $s_k$ corresponds to Step$k$ (Step1--Step5) defined by our template. 
If parsing fails (missing tags, wrong ordering, \etc), the extracted steps are marked invalid.

\noindent\textbf{Step-wise exact-match rewards.}
We compute strict step-wise rewards by exact matching against ground-truth step targets $\{s_k^*\}_{k=1}^5$.
For each step:
\begin{equation}
r_k(x,y) =
\mathbb{I}\big[\hat{s}_k = s_k^*\big] \in \{0,1\}.
\label{eq:rk}
\end{equation}


We define a format reward $w_{\text{fmt}}$ to encourage interpretable and well-structured outputs, which equals 1 if the output contains both the reasoning process $y^{\text{think}}$ and the final answer $y^{\text{ans}}$, and 0 otherwise. 
We aggregate three reward components into a scalar return:
\begin{equation}
R(x,y) = w_{\text{ans}}\, r_{\text{ans}}(x,y) +w_{\text{fmt}}\, r_{\text{fmt}}(x,y) + \sum_{k=1}^{5} w_k \, r_k(x,y),
\label{eq:allreward}
\end{equation}
where $r_{\text{ans}}(x,y)$ is the accuracy reward, which equals 1 if $y^{\text{ans}}$ matches the correct composition label, $w_{\text{ans}}$, $w_{\text{fmt}}$, and $w_k$ are weights for different rewards. 
Moreover, we assume that reasoning process constraints should be applied only when both the final prediction and the output format are correct. 
Therefore, $w_k > 0$ only if both $r_{\text{ans}}(x,y)$ and $r_{\text{fmt}}(x,y)$ equal 1.

\noindent\textbf{GRPO clipped objective with KL regularization.}
Given an input $x$, we sample a group of $G$ complete trajectories (\ie, $\mathcal{Y}_G(x)=\{y_1,\ldots,y_G\}$) from the current model $\pi_{\theta}$. 
For each trajectory $y_g$, we can compute a scalar reward $R_g$,
and we compute group statistics and normalize returns to obtain group-relative advantages:
\begin{equation}
A_g = (R_g - \mathrm{mean}({R_1, \dots, R_G}))/\mathrm{std}({R_1, \dots, R_G}).
\label{eq:group_adv}
\end{equation}
We broadcast the trajectory-level advantage to all decoding steps, \ie, $\hat{A}_{g,t} = A_g$. 
At each decoding step, the importance sampling ratio is defined as
\begin{equation}
\rho_{g,t}(\theta)=
\frac{\pi_{\theta}(y_{g,t}\mid x, y_{g,<t})}
{\pi_{\theta_{\mathrm{old}}}(y_{g,t}\mid x, y_{g,<t})}.
\label{eq:ratio}
\end{equation}
where $\pi_{\theta_{\mathrm{old}}}$ is the model before the current update. 
We optimize a PPO-style clipped surrogate objective with group-relative advantages:
\begin{equation}
\begin{adjustbox}{max width=\linewidth}
$\begin{aligned}
\mathcal{J}_{\mathrm{GRPO}}(\theta)
= &
\mathbb{E}_{x}\Bigg[\frac{1}{G}\sum_{g=1}^{G}\frac{1}{T_g}\sum_{t=1}^{T_g}\min\Big(\rho_{g,t}\hat{A}_{g,t}, \mathrm{clip}(\rho_{g,t},1-\delta,1+\delta)\hat{A}_{g,t} \Big) \\&-\beta\,\mathbb{D}_{\mathrm{KL}}\!\left(\pi_\theta\,\|\,\pi_{\mathrm{ref}}\right) \Bigg],
\end{aligned}$
\end{adjustbox}
\label{eq:grpo_obj}
\end{equation}
where $\delta$ is the clip range, $\pi_{\mathrm{ref}}$ is the model fixed after Stage I, and $\beta$ is the KL penalty coefficient. 
We compute the KL penalty as the average token-level log-probability difference along sampled trajectories:
\begin{equation}
\begin{adjustbox}{max width=\linewidth}
$\displaystyle
\mathbb{D}_{\mathrm{KL}}\!\left(\pi_\theta\,\|\,\pi_{\mathrm{ref}}\right)
=
\mathbb{E}_{y\sim \pi_{\theta}}\Bigg[
\frac{1}{T}\sum_{t=1}^{T}
\Big(
\log \pi_{\theta}(y_{t}\mid x,y_{<t}) 
-
\log \pi_{\mathrm{ref}}(y_{t}\mid x,y_{<t})
\Big)
\Bigg],
$
\end{adjustbox}
\label{eq:kl_def}
\end{equation}
where $T$ denotes the trajectory length. In practice, we maximize $\mathcal{J}_{\mathrm{GRPO}}(\theta)$ using minibatches of inputs.

We argue that this formulation is well-suited for policy exploration in our reasoning-based CZSL. 
After SFT, the model produces structured reasoning traces. 
GRPO then explores alternative reasoning paths under minimal constraints. 
In each iteration, it samples multiple strategies that may yield different predicted objects, and the reward guides learning toward paths that improve prediction accuracy.

\subsection{Inference Phase}
\label{sub:inference}



We adopt two evaluation settings to assess the reasoning performance of our PRPC.

\noindent\textit{Setting 1.} We require the model to produce a CoT output in the same format as used during training. 
We parse the output and extract the predicted attribute object label from $y^{\text{ans}}$, \ie, $\hat{c}$. 
A prediction is considered correct if it exactly matches the ground truth label $c$, \ie, $\text{Acc}(x)=\mathbb{I}\!\left[\hat{c} = c\right]$.


\noindent\textit{Setting 2.} Following the evaluation paradigm commonly used for classification with MLLMs~\cite{pratt2023does}, we prompt the model to generate an open-form text output. 
Specifically, the model describes the image using an attribute-object phrase $\hat{y}$, \eg, `a photo of \textit{attribute object}'. 
We then embed both the generated text $\hat{y}$ and the candidate label set (prefixed with `a photo of') using the CLIP text encoder. 
We select the label with the highest cosine similarity to $\hat{y}$ in the embedding space as the final predicted class $\hat{c}$. 
This setting aligns the inference procedure with the traditional closed-world classification protocol in CZSL and provides an additional measure of model performance.




\begin{table*}[htbp]
    \centering
    \resizebox{\textwidth}{!}{
        \begin{tabular}{lcccccccccccccccccccc}
        \toprule
        \multirow{2}{*}{\textbf{Models}} & \multicolumn{6}{c}{\textbf{MIT-States}} && \multicolumn{6}{c}{\textbf{C-GQA}} && \multicolumn{6}{c}{\textbf{VAW-CZSL}}\\
        \cline{2-7} \cline{9-14} \cline{16-21}
        & \rule{0pt}{2.3ex}AUC & HM & Seen & Unseen & Attr & Obj && AUC & HM & Seen & Unseen & Attr & Obj&&   AUC & HM & Seen & Unseen & Attr & Obj\\
        \hline
        CLIP \cite{CLIP} \rule{0pt}{2.3ex}& 11.0 & 26.1 & 30.2 & 46.0 & 33.2 & 51.1 && 1.4 & 8.6 & 7.5 & 25.0 & 13.9 & 30.2 && 0.2 & 3.7 & 3.9 & 9.6 & 7.7 & 24.0\\
        \hline
        DeepSeek-VL2-Tiny \cite{Deepseek-vl2} \rule{0pt}{2.3ex}& 4.7 & 16.6 & 17.0 & 16.2 & 21.8 & 34.9 && 0.6& 6.2 & 7.4 & 5.4 & 10.9 & 24.1 && 0.4 & 4.6 & 4.4 & 5.0 & 9.0 & 22.5\\
        LLaVA-Next-7B \cite{Llavanext} \rule{0pt}{2.3ex}& 4.7 & 16.3 & 21.6 & 13.0 & 18.4 & 36.8 && 3.3 & 15.5 & 16.6 & 14.6 & 23.5 & 33.5 && 0.7 & 7.0 & 7.1 & 6.9 & 11.2 & 25.7\\
        Molmo-7B-D \cite{molmo} \rule{0pt}{2.3ex}& 6.4 & 19.5 & 25.1 & 15.9 & 24.0 & 38.1 && 2.7 & 13.2 & 17.1 & 10.7 & 21.5 & 33.2 && 0.3 & 4.5 & 5.1 & 4.0 & 8.9 & 20.7\\
        InternVL2-8B \cite{InternVL} \rule{0pt}{2.3ex}& 6.8 & 20.7 & 19.9 & 21.6 & 27.9 & 40.2 && 3.9 & 15.8 & 15.4 & 16.3 & 21.9 & 32.6 && 0.6 & 6.2 & 7.4 & 5.3 & 10.9 & 24.1\\
        Qwen3.0-VL-8B \cite{Qwen} \rule{0pt}{2.3ex}& 10.9 & 26.1 & 25.2 & 27.1 & 33.0 & 46.6 && 5.2 & 20.2 & 21.8 & 18.9 & 27.7 & 36.5 && 0.6 & 6.1 & 5.8 & 6.5 & 9.9 & 23.1\\
        \hline
        \textbf{PRPC (Ours)} \rule{0pt}{2.3ex}& 11.5 & 29.2 & 27.9 & 30.7 & 36.5 & 49.8 && 7.3 & 23.5 & 24.2 & 22.7 & 30.6 & 40.5 && 1.4 & 9.3 & 10.2 & 8.6 & 13.6 & 31.8\\
        \bottomrule
        \end{tabular}
    }
    \caption{Results (\%) on MIT-States, C-GQA, and VAW-CZSL under standard CZSL similarity-based evaluation. We report Top-1 AUC, which balances between seen and unseen compositions with different bias terms. HM denotes the Harmonic Mean of Top-1 accuracies on seen images (Seen) and unseen images (Unseen). Best accuracy values of primitives \{Attr, Obj\} are also reported. }
    \label{tab:sota}
\end{table*}

\section{Experiments}

\subsection{Experimental Setup}
\noindent\textbf{Datasets.}
We evaluate our model on three benchmark datasets: 
1) \textit{MIT-States}~\cite{MIT} contains diverse real-world objects (\eg, cheese, sea) described by attributes (\eg, molten, dark). 
2) \textit{C-GQA}~\cite{CGE} is the most extensive CZSL dataset, which is newly created based on the Stanford GQA dataset \cite{StanfordGQA} for VQA tasks, composed of attributes (\eg, red, dirty) and objects (\eg, pen, window) commonly found in daily life. 
3) \textit{VAW-CZSL}~\cite{OADis} is a large-scale dataset derived from the VAW (Visual Attributes in the Wild) dataset \cite{VAW}, composed of attributes (\eg, furry, wet) and objects (\eg, dog, umbrella) grounded in real-world images. 
These three datasets contain diverse natural-scene images with long-tailed distributions, making them well-suited for evaluating zero-shot performance.

\noindent\textbf{Evaluation Metrics.} 
Traditional CZSL methods are typically formulated as a closed-set classification problem, where the model outputs comparable scores over all candidate compositions. 
In this setting, the widely used AUC metric evaluates the trade-off between accuracy on seen and unseen compositions by calibrating prediction bias through a tunable margin parameter~\cite{ICCV2019-TMN}. 
However, our approach reformulates CZSL as an open-form compositional reasoning task with MLLMs. 
The model directly generates a prediction without explicitly enumerating or scoring all logical candidates. 
Consequently, the calibration-based AUC metric is not directly applicable to our generative output. 
Therefore, we adopt distinct metrics for the two inference schemes introduced in Section~\ref{sub:inference}:
For \textit{Setting 1}, we report the Top-1 accuracy on seen images (Seen-$r$) and unseen images (Unseen-$r$), and the corresponding best harmonic mean (HM-$r$). Additionally, to analyze the reasoning chain, we evaluate the Top-1 accuracy for primitive components: Attribute (Attr-$r$), Object (Obj-$r$), and the full Composition (Pair-$r$). Note that strict string matching is used here. 
For \textit{Setting 2}, we follow the standard CZSL evaluation protocol and report the conventional metrics.

\subsection{Comparison with State-of-the-arts}

\noindent\textbf{Compared in the Standard CZSL Evaluation Protocol.}
In Table~\ref{tab:sota}, we compare our method with representative MLLMs and the CLIP baseline under the hybrid evaluation \textit{Setting~2}.
We adopt CLIP as a unified decision interface for all methods and evaluate them under the standard CZSL protocol. Our approach differs from the CLIP baseline only in how the text queries are generated, while sharing the same visual encoder, label space, and similarity-based inference, allowing us to isolate the effect of reasoning on compositional recognition.

Our PRPC framework employs an MLLM for progressive compositional reasoning, while the final prediction is obtained by projecting the generated composition through the CLIP text encoder into the closed-set label space.
Compared to direct MLLM baselines without structured reasoning, our method consistently improves recognition accuracy, indicating that intermediate verification and correction yield higher-quality compositional hypotheses for downstream projection.

When evaluated against CLIP under the same projection protocol, our approach achieves comparable performance and shows gains on several datasets, despite sharing the same CLIP text encoder.
This suggests that the improvements arise from enhanced reasoning and query generation quality rather than increased representation capacity.

More broadly, \textit{Setting~2} highlights the role of open-form reasoning even under closed-set evaluation.
While CLIP provides strong embedding alignment, it lacks explicit mechanisms to verify or revise object–attribute associations. In contrast, MLLM-based reasoning supports iterative hypothesis refinement before projection, offering a more robust and extensible paradigm for compositional recognition.

\begin{table}
    \centering
    \resizebox{0.49\textwidth}{!}{
        \begin{tabular}{cccccccccc}
        \toprule
        \multirow{2}{*}{\textbf{Datasets}} & \multicolumn{3}{c}{\textbf{Obj-$r$}} & \multicolumn{3}{c}{\textbf{Attr-$r$}}&\multirow{2}{*}{\textbf{Obj-$r$}$\uparrow$} &\multirow{2}{*}{\textbf{Attr-$r$}$\uparrow$}&\multirow{2}{*}{\textbf{Pair-$r$}}\\
        \cline{2-7}
        &S1&S3&S5&S2&S4&S5&&&\\
        \hline
        MIT-States& 35.1 & 40.8 & 41.0 & 21.4 & 22.2 & 22.4 & 5.9& 1.0 &12.7\\
        C-GQA& 47.9 & 49.8 & 50.2 & 34.7 & 35.5 & 35.6 & 2.3& 0.9 &24.5\\
        VAW-CZSL& 39.3 & 40.2 & 40.2 & 8.6 & 10.2 & 10.4 & 0.9& 1.8 &6.1\\
        \bottomrule
        \end{tabular}
    }
    \caption{Step-wise Results (\%) in the reasoning process.}
    \label{tab:sota-1}
\end{table}

\begin{table}
    \centering
    \resizebox{0.49\textwidth}{!}{
        \begin{tabular}{cccccccccc}
        \toprule
        \multirow{2}{*}{\textbf{Datasets}} & \multicolumn{3}{c}{\textbf{Obj-$r$}} & \multicolumn{3}{c}{\textbf{Attr-$r$}}&\multirow{2}{*}{\textbf{Obj-$r$}$\uparrow$} &\multirow{2}{*}{\textbf{Attr-$r$}$\uparrow$}&\multirow{2}{*}{\textbf{Pair-$r$}}\\
        \cline{2-7}
        &S1&S3&S5&S2&S4&S5&&&\\
        \hline
        MIT-States& 28.4 & 35.6 & 35.9 & 15.4 & 18.4 & 18.5 & 7.5& 3.1 & 11.0\\
        C-GQA& 41.3 & 46.6 & 47.8 & 28.3 & 31.2 & 31.2 & 6.5& 2.9 &22.4\\
        VAW-CZSL& 32.2 & 37.2 & 37.3 & 8.5 & 9.2 & 9.6 & 5.1& 1.1 & 5.9\\
        \bottomrule
        \end{tabular}
    }
    \caption{Step-wise Results (\%) in the reasoning process, with an incorrect object prediction manually enforced as the input at Step1.}
    \label{tab:sota-2}
\end{table}

\noindent\textbf{Results of Step-wise Reasoning and Error Correction.} 
We report the step-wise accuracies during reasoning under \textit{Setting 1} on the three datasets in Table \ref{tab:sota-1}.
PRPC achieves higher object and attribute prediction accuracy at middle stages, indicating that explicitly decomposing compositional recognition into iterative sequential decisions already yields more reliable primitive predictions. 
More importantly, our model attains high success rates in primitive correction during reasoning and yields substantial gains in final composition prediction performance.

\begin{table}
    \centering
    \resizebox{0.49\textwidth}{!}{
        \begin{tabular}{cccccccc}
        \toprule
        Datasets & Model & AUC & HM & Seen & Unseen & Attr & Obj\\
        \hline
        \multirow{4}{*}{MIT-States}& Qwen & 10.9 & 26.1 & 25.2 & 27.1 & 33.0 & 46.6 \\
        & Qwen+Stage I & 11.5 & 26.4 & 27.4 & 25.0 & 33.4 & 45.7\\
        & Qwen+Stage II & 10.8 & 24.3 & 24.4 & 24.2 & 32.9 & 45.1\\
        & PRPC & 11.5 & 29.2 & 27.9 & 30.7 & 36.5 & 49.8\\
        \hline
        \multirow{4}{*}{C-GQA}& Qwen & 5.2 &20.2 &21.8 &18.9 &27.7 &36.5\\
        & Qwen+Stage I & 6.2 & 21.9 & 23.5 & 20.6 & 28.9 & 39.6\\
        & Qwen+Stage II & 5.5 & 20.0 & 21.6 & 17.4 & 28.0 & 36.6\\
        & PRPC & 7.3 & 23.5 & 24.2 & 22.7 & 30.6 & 40.5 \\
        \hline
        \multirow{4}{*}{VAW-CZSL}& Qwen & 0.6 &6.12 &5.8 &6.5 &9.9 &23.1\\
        & Qwen+Stage I & 1.0 & 8.4 & 8.9 & 8.0 & 11.8 & 30.5\\
        & Qwen+Stage II & 0.7 & 6.0 & 6.6 & 5.5 & 9.1 & 23.4\\
        & PRPC & 1.4 & 9.3 & 10.2 & 8.6 & 13.6 & 31.8\\
        \bottomrule
        \end{tabular}
    }
    \caption{Results (\%) on three datasets with different training stages. }
    \label{tab:ab0}
\end{table}


To assess the robustness of the proposed reasoning framework, we conduct controlled error injection experiments by intentionally providing incorrect objects. 
Here, we use the CLIP text encoder to select the five objects most similar to each composition in the closed composition set.
For each sample, we introduce a randomly selected incorrect object into the question prompt, ensuring sufficient difficulty for target correction.
As the Table \ref{tab:sota-2} shows, PRPC recovers a portion of errors through mutual verification, indicating that the proposed correction loop is highly effective at revising early decisions.

These results highlight a key distinction between reasoning-based and similarity-based CZSL formulations. 
While CLIP-based methods produce a single static similarity score and cannot revise intermediate predictions, PRPC explicitly models compositional reasoning as an iterative decision process, enabling systematic error correction and improved compositional consistency. 
This capability is particularly important in scenarios where early predictions are uncertain or partially incorrect, and cannot be captured by existing embedding-based evaluation protocols.

\begin{table*}
    \centering
    \includegraphics[width=1.0\textwidth]{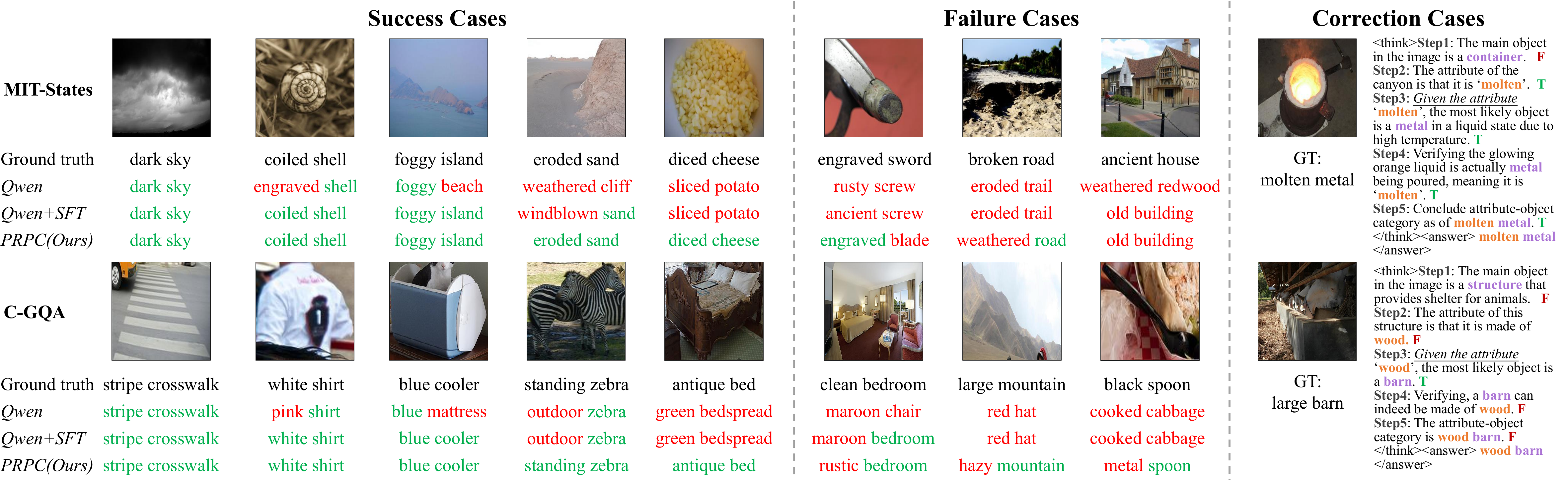}
    \caption{Qualitative comparison with Qwen3.0-VL. We present predictions on randomly selected test cases from MIT-States and C-GQA. Green/Red denotes the correct/wrong predictions. We also provide examples showing PRPC’s reasoning process with prediction correction.}
    \label{fig:qualitative}
\end{table*}

\subsection{Ablation Analysis}

\noindent\textbf{Effect of Two-Stage Training.}
We evaluate the effectiveness of two-stage training in our model, and report the results of ablation studies in Table \ref{tab:ab0}. 
We find that fine-tuning with Stage I alone yields a noticeable performance gain, whereas using Stage II alone leads to only marginal improvement. 
We think Stage I SFT is crucial because the subsequent GRPO relies on sparse exact-match rewards. 
Without SFT, early RL exploration often produces malformed trajectories (\eg, missing fields or an invalid step structure), leading to unreliable parsing and near-zero rewards. 
It slows optimization and encourages inefficient behaviors rather than reasoning.

\begin{table}
    \centering
    \resizebox{0.49\textwidth}{!}{
        \begin{tabular}{cccccccc}
        \toprule
        Datasets & Model & AUC & HM & Seen & Unseen & Attr & Obj\\
        \hline
        \multirow{4}{*}{MIT-States}& Qwen & 10.9 & 26.1 & 25.2 & 27.1 & 33.0 & 46.6  \\
        & PRPC w/o Steps1-5 & 10.8 & 26.3 & 25.1 & 27.6 & 32.0 & 45.6\\
        & PRPC w/o Steps3-5 & 10.9 & 26.7 & 25.2 & 28.5 & 33.3 & 45.5\\
        & PRPC & 11.5 & 26.4 & 27.4 & 25.0 & 33.4 & 45.7\\
        \hline
        \multirow{4}{*}{C-GQA}& Qwen & 5.2 & 20.2 & 21.8 & 18.9 & 27.7 & 36.5  \\
        & PRPC w/o Steps1-5 & 5.9 & 20.6 & 22.2 & 19.2 & 27.9 & 36.7\\
        & PRPC w/o Steps3-5 & 6.2 & 20.8 & 22.6 & 19.3 & 28.5 & 38.1\\
        & PRPC & 6.2 & 21.9 & 23.5 & 20.6 & 28.9 & 39.6\\
        \bottomrule
        \end{tabular}
    }
    \caption{Results (\%) with different CoT designs in Stage I SFT. None of these variants use Stage II training.}
    \label{tab:ab1}
\end{table}

\begin{table}
    \centering
    \resizebox{0.49\textwidth}{!}{
        \begin{tabular}{cccccccc}
        \toprule
        Datasets & Methods & AUC & HM & Seen & Unseen & Attr & Obj\\
        \hline
        \multirow{4}{*}{MIT-States}& No Step-wise Reward & 10.2 & 26.2 & 25.6 & 26.8 & 32.8 & 45.8 \\
        & Uniform Reward Across Steps & 10.4 & 26.1 & 25.4 & 26.7 & 32.5 & 46.3 \\
        & Higher Rewards in Early Steps & 11.1 & 28.1 & 26.7 & 29.5 & 35.2 & 48.6 \\
        & Higher Rewards in Later Steps & 11.5 & 29.2 & 27.9 & 30.7 & 36.5 & 49.8 \\
        \hline
        \multirow{4}{*}{C-GQA}& No Step-wise Reward & 6.8 & 22.7 & 24.6 & 21.2 & 30.3 & 40.4 \\
        & Uniform Reward Across Steps & 7.0 & 22.6 & 23.2 & 21.9 & 29.6 & 39.5 \\
        & Higher Rewards in Early Steps & 7.1 & 23.0 & 23.9 & 21.9 & 30.2 & 40.2 \\
        & Higher Rewards in Later Steps & 7.3 & 23.5 & 24.2 & 22.7 & 30.6 & 40.5 \\
        \bottomrule
        \end{tabular}
    }
    \caption{Results (\%) of different step-wise reward weights in Eq.\ref{eq:allreward}.}
    \label{tab:ab2}
\end{table}

\noindent\textbf{Effect of Bidirectional Corrective Reasoning in CoT.}
We evaluate the effectiveness of iterative reasoning design in CoT, and report the results of ablation studies in Table \ref{tab:ab1}. 
We additionally design two variants. 
In Table \ref{tab:ab1}, Line 2 fine-tunes using only the \texttt{<answer>} portion \(y^{\text{ans}}\) of the CoT, \ie, \(\alpha = 0\) in Eq. \(\ref{eq:sft}\); Line 3 fine-tunes after removing Steps3–5 from the \texttt{<think>} portion \(y^{\text{think}}\) in the ground-truth CoT. 
The results clearly demonstrate the effectiveness of using CoT for Bidirectional Corrective Reasoning, and in particular show that our proposed primitive-correction Steps3–5 provide a substantial performance boost.

\noindent\textbf{Effect of Different Step-Wise Rewards.}
We evaluate the effectiveness of different step-wise rewards, and report the results of ablation studies in Table \ref{tab:ab2}. 
Supervising only the final output or applying uniform rewards across steps leads to inferior performance, indicating that undifferentiated supervision is insufficient for multi-step reasoning. 
Emphasizing rewards on early steps improves initial predictions but degrades overall accuracy due to overconfident hypotheses. 
In contrast, assigning higher rewards to the verification and correction steps (Steps 3–4) consistently yields the best performance, highlighting the critical role of mutual refinement in compositional reasoning.

\subsection{Qualitative Results}

Figure~\ref{fig:qualitative} presents qualitative comparisons on MIT-States and C-GQA, and further illustrates the primitive prediction correction process through correction cases.
The vanilla Qwen model often produces visually plausible yet incorrect compositions, mainly due to confusion between fine-grained attributes or drifting toward correlated but invalid concepts (\eg, foggy beach, outdoor zebra). 
Supervised fine-tuning partially mitigates these issues by improving primitive recognition, but attribute–object alignment remains inconsistent, especially when attributes require subtle visual discrimination.
In contrast, the full PRPC model (\ie, including SFT+RL) consistently yields more accurate and coherent compositions. 
It successfully captures fine-grained attribute distinctions and recovers correct predictions in challenging cases such as `diced cheese' and `antique bed', where both Qwen and Qwen+SFT fail. 
Even in failure cases, our full PRPC produces predictions that are semantically closer to the ground truth, indicating reduced attribute hallucination and semantic drift. 
Overall, these qualitative results and correction cases highlight that the later steps (\ie, Steps3–4) play a critical role in refining compositional reasoning and enforcing semantic consistency.


\section{Conclusion}
\label{sec:conclusion}

In this paper, we revisit CZSL from a reasoning-centric perspective and formulate it as a structured multi-step inference problem. 
By enabling progressive verification and refinement between object and attribute predictions, our proposed approach PRPC improves compositional consistency and reduces error accumulation. 
Our results highlight reasoning-based learning as a complementary direction beyond similarity-based zero-shot compositional recognition.

\section*{Acknowledgments} 
This work was supported by the Beijing Natural Science - Xiaomi Innovation Joint Foundation (No. L253007).

\section*{Contribution Statement} 
Ziyi Chen and Haoyan Shi contributed equally to this work and are designated as co-first authors. Congyan Lang served as the corresponding author and is responsible for all communications related to this paper.

\bibliographystyle{named}
\bibliography{ijcai26}

\end{document}